\documentclass[final]{cvpr}

\usepackage{times}
\usepackage{epsfig}
\usepackage{graphicx}
\usepackage{amsmath}
\usepackage{amssymb}
\usepackage{cite}
\usepackage{multirow}
\usepackage{array}
\usepackage{colortbl}
\usepackage{algorithm}
\usepackage{algorithmicx}
\usepackage{algpseudocode}
\usepackage{comment}
\usepackage{color}
\usepackage{subfig}
\usepackage[table]{xcolor}

\usepackage[pagebackref=true,breaklinks=true,colorlinks,bookmarks=false]{hyperref}

\def\cvprPaperID{3191} 
\def\confYear{CVPR 2021}

\begin{document}

\title{PV-RAFT: Point-Voxel Correlation Fields for Scene Flow Estimation of Point Clouds}

\author{Yi Wei\textsuperscript{1,2,3}\thanks{Equal Contribution}, Ziyi Wang\textsuperscript{1,2,3}\footnotemark[1], Yongming Rao\textsuperscript{1,2,3}\footnotemark[1], Jiwen Lu\textsuperscript{1,2,3}\thanks{Corresponding author}, Jie Zhou\textsuperscript{1,2,3,4}\\
\textsuperscript{1}Department of Automation, Tsinghua University, China\\
\textsuperscript{2}State Key Lab of Intelligent Technologies and Systems, China\\
\textsuperscript{3}Beijing National Research Center for Information Science and Technology, China\\
\textsuperscript{4}Tsinghua Shenzhen International Graduate School, Tsinghua University, China \\
{\tt\small \{y-wei19, wziyi20\}@mails.tsinghua.edu.cn; raoyongming95@gmail.com;  \{lujiwen, jzhou\}@tsinghua.edu.cn} \\
}

\maketitle

\begin{abstract}
In this paper, we propose a Point-Voxel Recurrent All-Pairs Field Transforms (PV-RAFT) method to estimate scene flow from point clouds. Since point clouds are irregular and unordered, it is challenging to efficiently extract features from all-pairs fields in the 3D space, where all-pairs correlations play important roles in scene flow estimation. To tackle this problem, we present point-voxel correlation fields, which capture both local and long-range dependencies of point pairs. To capture point-based correlations, we adopt the K-Nearest Neighbors search that preserves fine-grained information in the local region. By voxelizing point clouds in a multi-scale manner, we construct pyramid correlation voxels to model long-range correspondences. Integrating these two types of correlations, our PV-RAFT makes use of all-pairs relations to handle both small and large displacements. We evaluate the proposed method on the  FlyingThings3D and KITTI Scene Flow 2015 datasets. Experimental results show that PV-RAFT outperforms state-of-the-art methods by remarkable margins.
\end{abstract}

\section{Introduction}

3D scene understanding~\cite{yang20203dssd,li2019stereo,shi2019pv,hou20193d,wang2018sgpn,su2018splatnet} has attracted more and more attention in recent years due to its wide real-world applications. As one fundamental 3D computer vision task, scene flow estimation~\cite{liu2019flownet3d,wupointpwc,puy2020flot,mittal2020just,gu2019hplflownet,hur2020self} focuses on computing the 3D motion field between two consecutive frames,  which provides important dynamic information. Conventionally, scene flow is directly estimated from RGB images \cite{mayer2016large,menze2015object,vedula2005three,vogel2013piecewise}. Since 3D data becomes easier to obtain, many works \cite{gu2019hplflownet,liu2019flownet3d,wupointpwc,puy2020flot} begin to focus on scene flow estimation of point clouds more recently.
\begin{figure}[tb]
	\centering
	\includegraphics[width=0.95\linewidth]{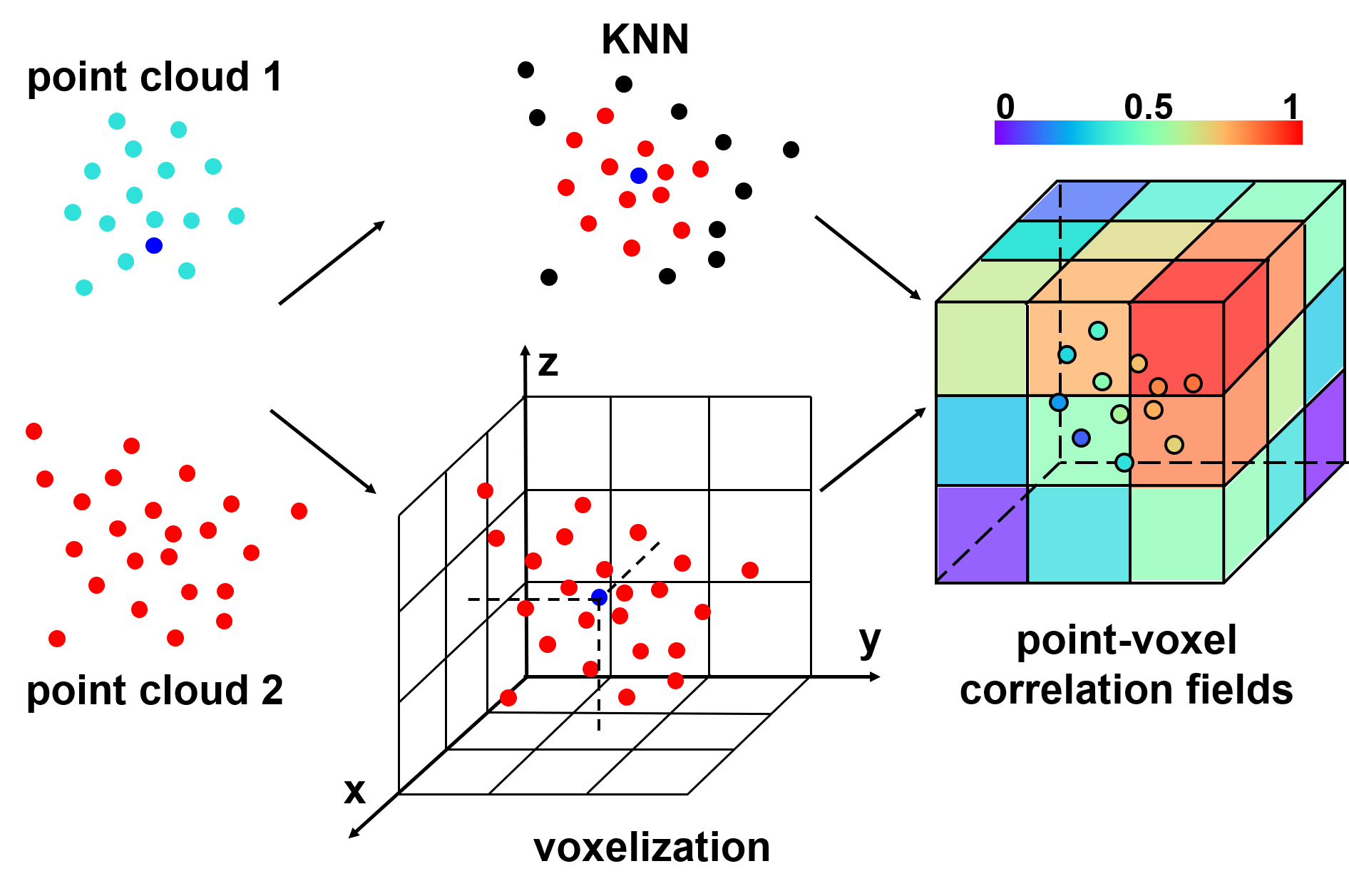}
	\caption{Illustration of the proposed point-voxel correlation fields. For a point in the source point cloud, we find its $k$-nearest neighbors in the target point cloud to extract point-based correlations. Moreover, we model long-range interactions by building voxels centered around this source point. Combining these two types of correlations, our PV-RAFT captures all-pairs dependencies to deal with both large and small displacements. }
	\label{fig:overview}
	\vspace{-3mm}
\end{figure}

Thanks to the recent advances in deep learning, many approaches adopt deep neural networks for scene flow estimation \cite{gu2019hplflownet,liu2019flownet3d,wupointpwc,puy2020flot,ushani2018feature}. Among these methods, \cite{liu2019flownet3d,wupointpwc} borrow ideas from \cite{ilg2017flownet,dosovitskiy2015flownet,sun2018pwc}, leveraging techniques in mature optical flow area. FlowNet3D designs a flow embedding module to calculate correlations between two frames. Built upon PWC-Net \cite{sun2018pwc}, PointPWC-Net \cite{wupointpwc} introduces a learnable point-based cost volume without the need of 4D dense tensors. These methods follow a coarse-to-fine strategy, where scene flow is first computed at low resolution and then upsampled to high resolution. However, this strategy has several limitations~\cite{teed2020raft} , \eg error accumulation from early steps and the tendency to miss fast-moving objects. One possible solution is to adopt Recurrent All-Pairs Field Transforms (RAFT) \cite{teed2020raft}, a state-of-the-art method for 2D optical flow, that builds correlation volumes for all pairs of pixels. Compared with the coarse-to-fine strategy, the all-pairs field preserves both local correlations and long-range relations. Nevertheless, it is non-trivial to lift it to the 3D space. Due to the irregularity of point clouds, building structured all-pairs correlation fields becomes challenging. Moreover, since point clouds are unordered, it is difficult to efficiently look up neighboring points of a 3D position. Unfortunately, the correlation volumes used in previous methods \cite{gu2019hplflownet,liu2019flownet3d,wupointpwc} only consider near neighbors, which fails to capture all-pairs relations.

To address these issues, we present point-voxel correlation fields that aggregate the advantages of both point-based and voxel-based correlations (illustrated in Figure \ref{fig:overview}). As mentioned in \cite{shi2019pv, tang2020searching,liu2019point}, point-based features maintain fine-grained information while voxel-based operation efficiently encodes large point set. Motivated by this fact, we adopt K-Nearest Neighbor (KNN) search to find a fixed number of neighboring points for point-based correlation fields. Meanwhile, we voxelize target point clouds in a multi-scale fashion to build pyramid correlation voxels. These voxel-based correlation fields collect long-term dependencies and guide the predicted direction. Moreover, to save memory, we present a truncation mechanism to abandon the correlations with low scores.

Based on point-voxel correlation fields, we propose a Point-Voxel Recurrent All-Pairs Field Transforms (PV-RAFT) method to construct a new network architecture for scene flow estimation of point clouds. Our method first employs a feature encoder to extract per-point features, which are utilized to build all-pair correlation fields. Then we adopt a GRU-based operator to update scene flow in an iterative manner, where we leverage both point-based and voxel-based mechanisms to look up correlation features. Finally, a refinement module is introduced to smooth the estimated scene flow. To evaluate our method, we conducted extensive experiments on the FlyingThings3D \cite{mayer2016large} and KITTI \cite{menze2015object,menze2015joint} datasets. Results show that our PV-RAFT outperforms state-of-the-art methods by a large margin. The code is available at {\small \url{https://github.com/weiyithu/PV-RAFT}}.

\section{Related Work}
\noindent \textbf{3D Deep Learning:}
Increased attention has been paid to 3D deep learning \cite{wei2019conditional,jiang2020pointgroup,rao2020global,shi2019pointrcnn,qi2017pointnet,qi2017pointnet++,yang20203dssd,li2019stereo,shi2019pv,hou20193d,wang2018sgpn,qi2019deep} due to its wide applications. As a pioneer work, PointNet \cite{qi2017pointnet} is the first deep learning framework directly operating on point clouds. It uses a max pooling layer to aggregate features of unordered set. PointNet++ \cite{qi2017pointnet++} introduces a hierarchical structure by using PointNet as a unit module. Kd-network \cite{klokov2017escape} equips a kd-tree to divide point clouds and compute a sequence of hierarchical representations. DGCNN \cite{wang2019dynamic} models point clouds as a graph and utilizes graph neural networks to extract features. Thanks to  these architectures, great achievements have been made in many 3D areas, \eg 3D recognition \cite{li2018pointcnn,li2018so,qi2017pointnet,qi2017pointnet++},  3D segmentation \cite{jiang2020pointgroup,hou20193d,wang2018sgpn}. Recently, several works \cite{shi2019pv,tang2020searching,liu2019point} simultaneously leverage point-based and voxel-based methods to operate on point clouds. Liu \etal \cite{liu2019point} present Point-Voxel CNN (PVCNN) for efficient 3D deep learning. It combines voxel-based CNN and point-based MLP to extract features. As a follow-up, Tang \etal \cite{tang2020searching} design SPVConv \cite{tang2020searching} which adopts Sparse Convolution with the high-resolution point-based network. They further propose 3D-NAS to search the best architecture. PV-RCNN \cite{shi2019pv} takes advantage of high-quality 3D proposals from 3D voxel CNN and accurate location information from PointNet-based set abstraction operation. Instead of equipping point-voxel architecture to extract features, we design point-voxel correlation fields to capture correlations.

\noindent \textbf{Optical Flow Estimation:} Optical flow estimation \cite{ilg2017flownet,dosovitskiy2015flownet,ranjan2017optical,hui2018liteflownet,truong2020glu} is a hot topic in 2D area. FlowNet \cite{dosovitskiy2015flownet} is the first trainable CNN for optical flow estimation, adopting a U-Net autoencoder architecture. Based on \cite{dosovitskiy2015flownet}, FlowNet2 \cite{ilg2017flownet} stacks several FlowNet models to compute large-displacement optical flows. With this cascaded backbone, FlowNet2 \cite{ilg2017flownet} outperforms FlowNet \cite{dosovitskiy2015flownet} by a large margin. To deal with large motions, SPyNet \cite{ranjan2017optical} adopts the coarse-to-fine strategy with a spatial pyramid. Beyond SPyNet \cite{ranjan2017optical},  PWC-Net \cite{sun2018pwc}  builds a cost volume by limiting the search range at each pyramid level. Similar to PWC-Net,  LiteFlowNet \cite{hui2018liteflownet} also utilizes multiple correlation layers operating on a feature pyramid. Recently, GLU-Net \cite{truong2020glu} combines global and local correlation layers with  an adaptive resolution strategy, which  achieves both high accuracy and robustness. Different from the coarse-to-fine strategy, RAFT \cite{teed2020raft} constructs the multi-scale 4D correlation volume for all pairs of pixels. It further updates the flow field through a recurrent unit iteratively, and achieves state-of-the-art performance on optical flow estimation task. The basic structure of our PV-RAFT is similar to theirs. However, we adjust the framework to  fit point clouds data format and propose point-voxel correlation fields to leverage all-pairs relations.

\noindent \textbf{Scene Flow Estimation:}
First introduced in \cite{vedula2005three}, scene flow is the three-dimension vector to describe the motion in real scenes.  Beyond this pioneer work, many studies estimate scene flow from RGB images \cite{huguet2007variational,pons2007multi,wedel2008efficient,wedel2011stereoscopic,vcech2011scene,vogel20113d,vogel2013piecewise,vogel20153d,basha2013multi}. Based on stereo sequences, \cite{huguet2007variational} proposes a variational method to estimate scene flow. Similar to  \cite{huguet2007variational}, \cite{wedel2008efficient} decouples the position and velocity estimation steps with consistent displacements in the stereo images. \cite{vogel20153d} represents dynamic scenes as a collection of rigidly moving planes and accordingly introduces a piecewise rigid scene model. With the development of 3D sensors, it becomes easier to get high-quality 3D data. More and more works focus on how to leverage point clouds for scene flow estimation \cite{dewan2016rigid,ushani2017learning,ushani2018feature,gu2019hplflownet,liu2019flownet3d,wupointpwc,puy2020flot}.  FlowNet3D \cite{liu2019flownet3d} introduces two layers to simultaneously learn deep hierarchical features of point clouds and flow embeddings. Inspired by Bilateral Convolutional Layers, HPLFlowNet \cite{gu2019hplflownet} projects unstructured point clouds onto a permutohedral lattice. Operating on  permutohedral lattice points, it can efficiently calculate scene flow. Benefiting from the coarse-to-fine strategy, PointPWC-Net \cite{wupointpwc} proposes cost volume, upsampling, and warping layers for scene flow estimation. Different from the above methods, FLOT \cite{puy2020flot} adopts the optimal transport to find correspondences. However, the correlation layers introduced in these methods only consider the neighbors in a local region, which fail to efficiently capture long-term dependencies. With point-voxel correlation fields, our PV-RAFT captures both local and long-range correlations.


\begin{figure*}[tb]
	\centering
	\includegraphics[width=0.95\linewidth]{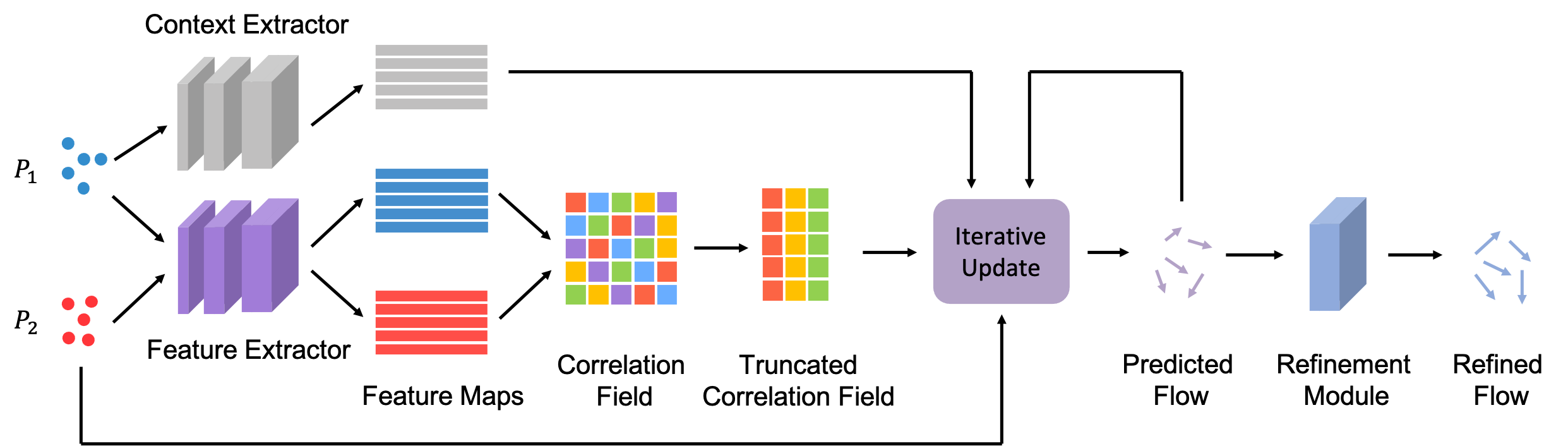}
	\caption{Illustration of the proposed PV-RAFT architecture. The feature extractor encodes high dimensional features of both $P_1$ and $P_2$, while the context extractor only encodes context features of $P_1$. We  calculate the matrix dot product of two feature maps to construct all-pair correlation fields. The truncated correlation field is then used in iterative update block to save memory. The detailed structure of 'Iterative Update' module can be found in Figure \ref{fig:iteration}. The predicted flow from the iteration block finally converges to a static status and is fed into the separately trained refinement module. We use the refined flow as the final scene flow prediction.}
	\label{fig:pipeline}
	\vspace{-3mm}
\end{figure*}

\section{Approach} \label{approach}
To build all-pairs fields, it is important to design a correlation volume which can capture both short-range and long-range relations. In this section, we first explain how to construct point-voxel correlation fields on point clouds. Then we will introduce the pipeline of our Point-Voxel Recurrent All-Pairs Field Transforms (PV-RAFT).

\subsection{Point-Voxel Correlation Fields} \label{sec:corr}
We first construct a full correlation volume based on feature similarities between all pairs. Given point clouds features $E_\theta(P_1)\in \mathbb{R}^{N_1\times D}, E_\theta(P_2)\in \mathbb{R}^{N_2\times D}$, where $D$ is the feature dimension, the correlation fields $\mathbf{C}\in \mathbb{R}^{N_1\times N_2}$ can be easily calculated by matrix dot product:
\begin{equation}
\mathbf{C} = E_\theta(P_1) \cdot E_\theta(P_2)
\end{equation}

\noindent \textbf{Correlation Lookup:}
The correlation volume $\mathbf{C}$ is built only once and is kept as a lookup table for flow estimations in different steps. Given a source point  $p_1 = (x_1, y_1, z_1) \in P_1$, a target point $p_2 = (x_2, y_2, z_2) \in P_2$ and an estimated scene flow $f = (f_1, f_2, f_3) \in \textbf{f}$, the source point is expected to move to $q = (x_1 + f_1, x_2 + f_2, x_3 + f_3) \in Q$, where $Q$ is the translated point cloud. We can easily get the correlation fields between $Q$ and $P_2$ by searching the neighbors of $Q$ in $P_2$ and looking up the corresponding correlation values in $\mathbf{C}$. Such looking-up procedure avoids extracting features of $Q$ and calculating matrix dot product repeatedly while keeping the all-pairs correlations available at the same time. Since 3D points data is not structured in the dense voxel, grid sampling is no longer useful and we cannot directly convert 2D method \cite{teed2020raft} into 3D version. Thus, the main challenge is how to locate neighbors and look up correlation values efficiently in the 3D space.

\noindent \textbf{Truncated Correlation:}
According to our experimental results, not all correlation entries are useful in the subsequent correlation lookup process. The pairs with higher similarity often guide the correct direction of flow estimation, while dissimilar pairs tend to make little contribution. To save memory and increase calculation efficiency in correlation lookup, for each point in $P_1$, we select its top-$M$ highest correlations. Specifically, we will get truncated correlation fields  $\mathbf{C}_M\in \mathbb{R}^{N_1\times M}$, where $M < N_2$ is the pre-defined truncation number. The point branch and voxel branch are built upon truncated correlation fields.

\noindent \textbf{Point Branch:}
A common practice to locate neighbors in 3D point clouds is to use K-Nearest Neighbors (KNN) algorithm. Suppose the  top-k nearest neighbors of $Q$ in $P_2$ is $\mathcal{N}_k = \mathcal{N}(Q)_k$ and their corresponding correlation values are $\textbf{C}_M(\mathcal{N}_k)$, the correlation feature between $Q$ and $P_2$ can be defined as:
\begin{equation}
\textbf{C}_p(Q, P_2) = \max_k({\rm MLP}({\rm concat}(\textbf{C}_M(\mathcal{N}_k), \mathcal{N}_k - Q)))
\end{equation}
where \textit{concat} stands for concatenation and $\max$ indicates a max pooling operation on $k$ dimension. We briefly note $\mathcal{N}(Q)$ as $\mathcal{N}$ in the following statements as all neighbors are based on $Q$ in this paper. The point branch extracts fine-grained correlation features of the estimated flow since the nearest neighbors are often close to the query point, illustrated in the upper branch of Figure \ref{fig:overview}. While the point branch is able to capture local correlations, long-range relations are often not taken into account in KNN scenario. Existing methods try to solve this problem by implementing the coarse-to-fine strategy, but error often accumulates if estimates in the coarse stage are not accurate.

\noindent \textbf{Voxel Branch:}
To tackle the problem mentioned above, we propose a voxel branch to capture long-range correlation features. Instead of voxelizing $Q$ directly, we build voxel neighbor cubes centered around $Q$ and check which points in $P_2$ lie in these cubes. Moreover, we also need to know each point's relative direction to $Q$. Therefore, if we denote sub-cube side length by $r$ and cube resolution by $a$, then the neighbor cube of $Q$ would be a $a\times a\times a$ Rubik's cube:
\begin{align}
\mathcal{N}_{r,a} &= \{\mathcal{N}_r^{(\mathbf{i})} \lvert \mathbf{i} \in \mathbb{Z}^3\} \\
\mathcal{N}_r^{(\mathbf{i})} &= \{Q + \mathbf{i} * r + \mathbf{dr}  \lvert \, \lVert{\mathbf{dr}}\rVert_1 \leq \frac{r}{2}\}
\end{align}
where $\mathbf{i} = [i, j, k]^T, \lceil{-\frac{a}{2}}\rceil \leq i,j,k \leq \lceil{\frac{a}{2}}\rceil \in \mathbb{Z}$ and each $r\times r\times r$ sub-cube $\mathcal{N}_r^{(\mathbf{i})}$ indicates a specific direction of neighbor points (\eg, $[0, 0, 0]^T$ indicates the central sub-cube). Then we identify all neighbor points in the sub-cube $\mathcal{N}_r^{(\mathbf{i})}$ and average their correlation values to get sub-cube features. The correlation feature between $Q$ and $P_2$ can be defined as:
\begin{equation}
\textbf{C}_v(Q, P_2) = {\rm MLP}\left( \mathop{\rm concat}\limits_{\mathbf{i}}\left(\frac{1}{n_{\mathbf{i}}}\sum_{n_{\mathbf{i}}}\textbf{C}_M\left(\mathcal{N}_r^{(\mathbf{i})}\right)\right)\right)
\end{equation}
where $n_{\mathbf{i}}$ is the number of points in $P_2$ that lie in the $\mathbf{i}^{th}$ sub-cube of $Q$ and $\textbf{C}_v(Q, P_2) \in \mathbb{R}^{N_1\times a^3}$. Please refer to the lower branch of Figure \ref{fig:overview} for illustration.

The Voxel branch helps to capture long-range correlation features as $r, a$ could be large enough to cover distant points. Moreover, we propose to extract pyramid correlation voxels with fixed cube resolution $a$ and proportionate growing sub-cube side length $r$. During each pyramid iteration, $r$ is doubled so that the neighbor cube expands to include farther points. The pyramid features are concatenated together before feeding into the MLP layer.

\subsection{PV-RAFT}
Given the proposed correlation fields that combine the fine-grained and long-range features, we build a deep neural network for scene flow estimation. The pipeline consists of four stages: (1) feature extraction, (2) correlation fields construction, (3) iterative scene flow estimation, (4) flow refinement. The first three stages are differentiable in an end-to-end manner, while the fourth one is trained separately with previous parts frozen. Our framework is called PV-RAFT and in this section we will introduce it in detail. Please refer to Figure \ref{fig:pipeline} for illustration.

\noindent \textbf{Feature Extraction:}
The feature extractor $E_\theta$ encodes point clouds with mere coordinates information into higher dimensional feature space, as $E_\theta: \mathbb{R}^{n \times 3} \mapsto \mathbb{R}^{n \times D}$. Our backbone framework is based on PointNet++ \cite{qi2017pointnet++}. For consecutive point clouds input $P_1, P_2$, the feature extractor outputs $E_\theta (P_1), E_\theta (P_2)$ as backbone features. Besides, we design a content feature extractor $E_\gamma$ to encode context feature of $P_1$. Its structure is exactly the same as feature extractor $E_\theta$, without weight sharing. The output context feature $E_\gamma (P_1)$ is used as auxiliary context information in GRU iteration.

\noindent \textbf{Correlation Fields Construction:}
As is introduced in Section \ref{sec:corr}, we build all-pair correlation fields $\mathbf{C}$ based on backbone features $E_\theta (P_1), E_\theta (P_2)$. Then we truncate it according to correlation value sorting and keep it as a lookup table for later iterative updates.

\begin{figure*}[tb]
	\centering
	\includegraphics[width=0.95\linewidth]{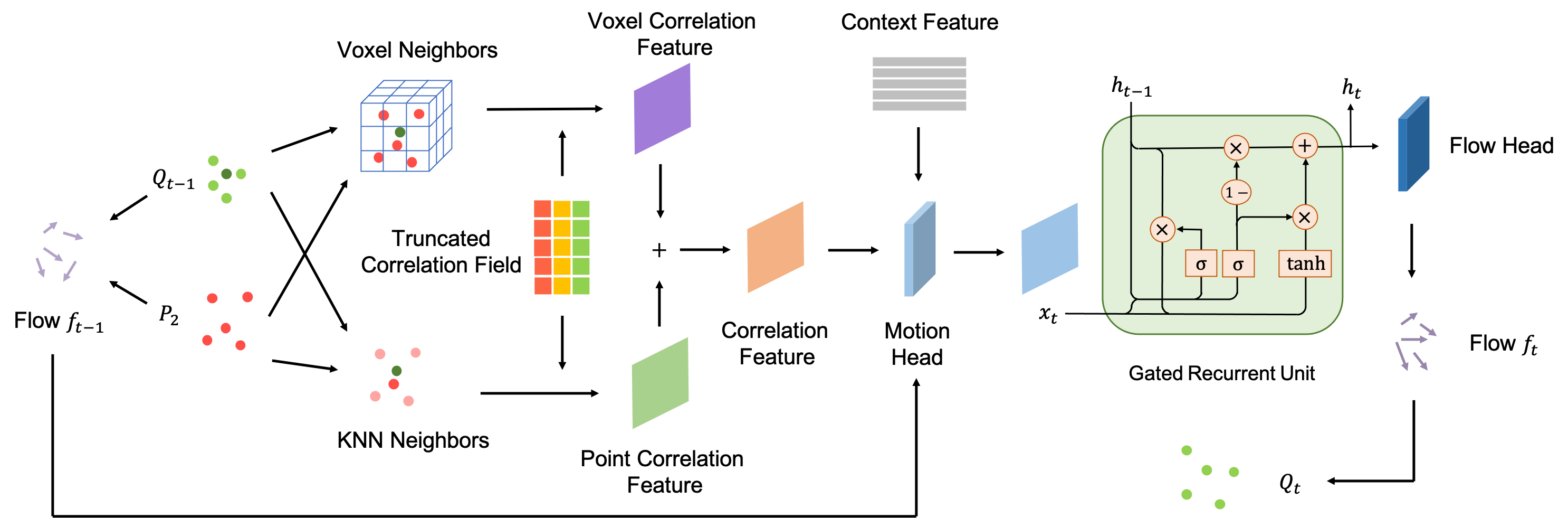}
	\caption{Illustration of the iterative update. This figure is a detailed explanation of the 'Iterative Update' module in Figure \ref{fig:pipeline}. During iteration $t$, we find both voxel neighbors and KNN  of $Q_{t-1}$ in $P_2$. This helps us extract long-range voxel correlation features and fine-grained point correlation features from the truncated correlation field. The combined correlation feature, together with context feature and current flow estimate $f_{t-1}$ are fed to a convolutional motion head. The output is used as $x_t$ of the Gated Recurrent Unit (GRU). Finally, the flow head encodes the hidden state $h_t$ of GRU to predict the residual of flow estimation, which is used to update $f_t$ and $Q_t$.}
	\label{fig:iteration}
	\vspace{-3mm}
\end{figure*}

\noindent \textbf{Iterative Flow Estimation:}
The iterative flow estimation begins with the initialize state $\mathbf{f}_0=0$. With each iteration, the scene flow estimation is updated upon the current state: $\mathbf{f}_{t+1}=\mathbf{f}_t + \Delta\mathbf{f}$. Eventually, the sequence converges to the final prediction $\mathbf{f_T}\to \textbf{f}^*$. Each iteration takes the following variables as input: (1) correlation features, (2) current flow estimate, (3) hidden states from the previous iteration, (4) context features. First, the correlation features are the combination of both fine-grained point-based ones and long-range pyramid-voxel-based ones:
\begin{equation}
\mathbf{C}_t = \mathbf{C}_p(Q_t, P_2) + \mathbf{C}_v(Q_t, P_2)
\end{equation}
Second, the current flow estimation is simply the direction vector between $Q_t$ and $P_1$:
\begin{equation}
\mathbf{f}_t = Q_t - P_1
\end{equation}
Third, the hidden state $h_t$ is calculated by GRU cell\cite{teed2020raft}:
\begin{align}
&z_t = \sigma(\text{Conv}_{\text{1d}}([h_{t-1}, x_t], W_z)) \\
&r_t = \sigma(\text{Conv}_{\text{1d}}([h_{t-1}, x_t], W_r)) \\
&\hat{h_t} = \tanh(\text{Conv}_{\text{1d}}([r_t \odot h_{t-1}, x_t], W_h)) \\
& h_{t} = (1 - z_t) \odot h_{t-1} + z_t \odot \hat{h_t}
\end{align}
where $x_t$ is a concatenation of correlation $\mathbf{C}_t$, current flow $\mathbf{f}_t$ and context features $E_\gamma (P_1)$. Finally, the hidden state $h_t$ is fed into a small convolutional network to get the final scene flow estimate $\textbf{f}^*$. The detailed iterative update process is illustrated in Figure \ref{fig:iteration}.

\noindent \textbf{Flow Refinement:}
The purpose of designing this flow refinement module is to make scene flow prediction $\textbf{f}^*$ smoother in the 3D space. Specifically, the estimated scene flow from previous stages is fed into three convolutional layers and one fully connected layer. To update flow for more iterations without out of memory, the refinement module is not trained end-to-end with other modules. We first train the backbone and iterative update module, then we freeze the weights and train the refinement module alone.

\subsection{Loss Function}
\noindent \textbf{Flow Supervision:}
We follow the common practice of supervised scene flow learning to design our loss function. In detail, we use $l_1$-norm between the ground truth flow $\mathbf{f}_{gt}$ and estimated flow $\mathbf{f}_{est}$ for each iteration:
\begin{equation}
\mathcal{L}_{iter} = \sum_{t=1}^{T}{w^{(t)}\lVert{(\mathbf{f}_{est}^{(t)} - \mathbf{f}_{gt})}\rVert_1}
\end{equation}
where $T$ is the total amount of iterative updates, $\mathbf{f}_{est}^{(t)}$ is the flow estimate at $t^{th}$ iteration, and $w^{(t)}$ is the weight of
$t^{th}$ iteration:
\begin{equation}
w^{(t)} = \gamma * (T - t - 1)
\end{equation}
where $\gamma$ is a hyper-parameter and we set $\gamma = 0.8$ in our experiments.

\noindent \textbf{Refinement Supervision:}
When we freeze the weights of previous stages and only train the refinement module, we design a similar refinement loss:
\begin{align}
\mathcal{L}_{ref} &= \lVert{(\mathbf{f}_{ref} - \mathbf{f}_{gt})}\rVert_1
\end{align}
where $\mathbf{f}_{ref}$ is the flow prediction from refinement module.


\section{Experiments}

In this section, we conducted extensive experiments to verify the superiority of our PV-RAFT. We first introduce the experimental setup, including datasets, implementation details and evaluation metrics. Then we show main results on the FlyingThings3D \cite{mayer2016large} and KITTI \cite{menze2015object,menze2015joint} datasets, as well as ablation studies. Finally, we give a further analysis of PV-RAFT to better illustrate the effectiveness of our proposed method.

\begin{table*}[tb]\footnotesize
	\centering
	\caption{Performance comparison on the FlyingThings3D and KITTI datasets. All methods are trained on FlyingThings3D in a supervised manner. The best results for each dataset are marked in bold.}
	\resizebox{0.8\textwidth}{!}{
		\begin{tabular}{l|l|cccc}
			\hline
			Dataset &Method  &EPE(m)$\downarrow$           & Acc Strict$\uparrow$          & Acc Relax$\uparrow$          & Outliers$\downarrow$      \\ \hline
			\multirow{5}{*}{FlyingThings3D}&
			FlowNet3D~\cite{liu2019flownet3d}
			& 0.1136    & 0.4125    & 0.7706    & 0.6016       \\
			&HPLFlowNet~\cite{gu2019hplflownet}
			& 0.0804    & 0.6144    & 0.8555    & 0.4287       \\
			&PointPWC-Net~\cite{wupointpwc}
			& 0.0588    & 0.7379    & 0.9276    & 0.3424       \\
			&FLOT~\cite{puy2020flot}
			& 0.052     & 0.732     & 0.927     & 0.357        \\
			&PV-RAFT
			&\bf{0.0461}&\bf{0.8169}&\bf{0.9574}&\bf{0.2924}       \\ \hline
			\multirow{5}{*}{KITTI}
			&FlowNet3D~\cite{liu2019flownet3d}
			& 0.1767    & 0.3738    & 0.6677    & 0.5271       \\
			&HPLFlowNet~\cite{gu2019hplflownet}
			& 0.1169    & 0.4783    & 0.7776    & 0.4103       \\
			&PointPWC-Net~\cite{wupointpwc}
			& 0.0694    & 0.7281    & 0.8884    & 0.2648       \\
			&FLOT~\cite{puy2020flot}
			&\bf{0.056} & 0.755     & 0.908     & 0.242        \\
			&PV-RAFT
			& \bf{0.0560}&\bf{0.8226}&\bf{0.9372}&\bf{0.2163}       \\ \hline
		\end{tabular}
		
	}
	\label{tab:main}
\end{table*}

\subsection{Experimental Setup}
\noindent \textbf{Datasets:}
Same with \cite{gu2019hplflownet,liu2019flownet3d,wupointpwc,puy2020flot}, we trained our model on the FlyingThings3D \cite{mayer2016large} dataset and evaluated it on both FlyingThings3D \cite{mayer2016large} and KITTI \cite{menze2015object,menze2015joint} datasets. We followed \cite{gu2019hplflownet} to preprocess data.  As a large-scale synthetic dataset, FlyingThings3D is the first benchmark for scene flow estimation.  With the objects from ShapeNet \cite{chang2015shapenet}, FlyingThings3D consists of rendered stereo and RGB-D images. Totally, there are 19,640 pairs of samples in the training set and 3,824 pairs in the test set. Besides, we kept aside 2000 samples from the training set for validation. We lifted depth images to point clouds and optical flow to scene flow instead of operating on RGB images. As another benchmark, KITTI Scene Flow 2015 is a dataset for scene flow estimation in real scans \cite{menze2015object,menze2015joint}. It is built from KITTI raw data by annotating dynamic motions. Following previous works \cite{gu2019hplflownet,liu2019flownet3d,wupointpwc,puy2020flot}, we evaluated on 142 samples in the training set since point clouds were not available in the test set. Ground points  were removed by height (0.3m). Further, we deleted points whose depths are larger than 35m.

\noindent \textbf{Implementation Details:}
We randomly sampled 8192 points in each point cloud to train PV-RAFT. For the point branch, we searched 32 nearest neighbors. For the voxel branch, we set cube resolution $a=3$ and built 3-level pyramid with $r=0.25,0.5,1$. To save memory, we set truncation number $M$ as 512. We updated scene flow for 8 iterations during training and evaluated the model with 32 flow updates. The backbone and iterative module were trained for 20 epochs. Then, we fixed their weights with 32 iterations and trained the refinement module for another 10 epochs. PV-RAFT was implemented in PyTorch \cite{paszke2017automatic}. We utilized Adam optimizer \cite{kingma2014adam} with initial learning rate as 0.001 .

\noindent \textbf{Evaluation Metrics:}
We adopted four evaluation metrics used in \cite{gu2019hplflownet,liu2019flownet3d,wupointpwc,puy2020flot}, including EPE, Acc Strict, Acc Relax and Outliers. We denote estimated scene flow and ground-truth scene flow as $f_{est}$ and $f_{gt}$ respectively. The evaluation metrics are defined as follows:

\noindent $\bullet$ \textbf{EPE}: $||f_{est}-f_{gt}||_2$. The end point error averaged on each point in meters.

\noindent $\bullet$ \textbf{Acc Strict}: the percentage of points whose \textbf{EPE} $< 0.05m$ or relative error $<5\%$.

\noindent $\bullet$ \textbf{Acc Relax}: the percentage of points whose \textbf{EPE} $< 0.1m$ or relative error $<10\%$.

\noindent $\bullet$ \textbf{Outliers}: the percentage of points whose \textbf{EPE} $>0.3m$ or relative error $>10\%$.

\begin{figure*}[tb]
	\centering
	\includegraphics[width=1.0\linewidth]{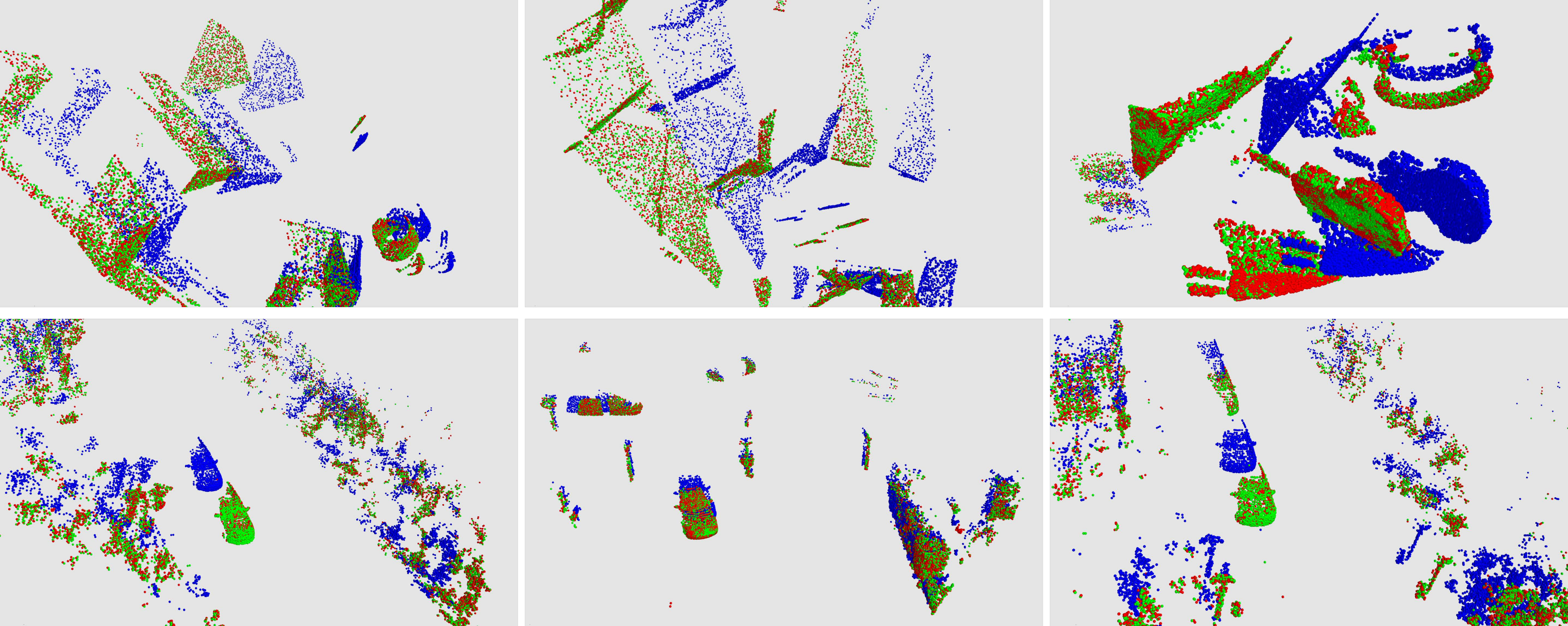}
	\caption{Qualitative results on FlyingThings3D (top) and KITTI (bottom). Blue points and red points indicate $P_1$ and $P_2$ respectively. Translated points $P_1$ + $\textbf{f}$ are in green. Our PV-RAFT can deal with both small and large displacements' cases.}
	\label{fig:visual}
	\vspace{-1mm}
\end{figure*}

\subsection{Main Results}
Quantitative results on the FlyingThings3D and KITTI datasets are shown in Table \ref{tab:main}. Our PV-RAFT achieves state-of-the-art performances on both datasets, which verifies its superiority and generalization ability. Especially, for Outliers metric, our method outperforms FLOT by 18.1\% and 10.6\% on two datasets respectively. The qualitative results in Figure \ref{fig:visual} further demonstrate the effectiveness of PV-RAFT. The first row and second row present visualizations on the FlyingThings3D and KITTI datasets respectively. As we can see, benefiting from point-voxel correlation fields, our method can accurately predict
both small and large displacements.

\begin{table*}[tb]\footnotesize
	\centering
	\caption{Ablation Studies of PV-RAFT on the FlyingThings3D dataset. We incrementally applied point-based correlation, voxel-based correlation and refinement module to the framework.}
	\resizebox{0.8\textwidth}{!}{
		\begin{tabular}{ccc|cccc}
			\hline
			point-based &voxel-based&refine &\multirow{2}{*}{EPE(m)$\downarrow$}           & \multirow{2}{*}{Acc Strict$\uparrow$}          & \multirow{2}{*}{Acc Relax$\uparrow$}           & \multirow{2}{*}{Outliers$\downarrow$}      \\
			correlation & correlation & module & & & & \\
			\hline
			$\checkmark$&  &
			& 0.0741    & 0.6111    & 0.8868    & 0.4549       \\
			& $\checkmark$ &
			&  0.0712	&0.6146	    &0.8983	    &0.4492      \\
			$\checkmark$ & $\checkmark$ &
			& 0.0534    & 0.7348    & 0.9418    & 0.3645       \\
			$\checkmark$ & $\checkmark$ &   $\checkmark$
			&\bf{0.0461}&\bf{0.8169}&\bf{0.9574}&\bf{0.2924}        \\
			 \hline
		\end{tabular}
		
	}
	\label{tab:ablation}
\end{table*}

\begin{figure*}[tb]
	\centering
	\includegraphics[width=1\linewidth]{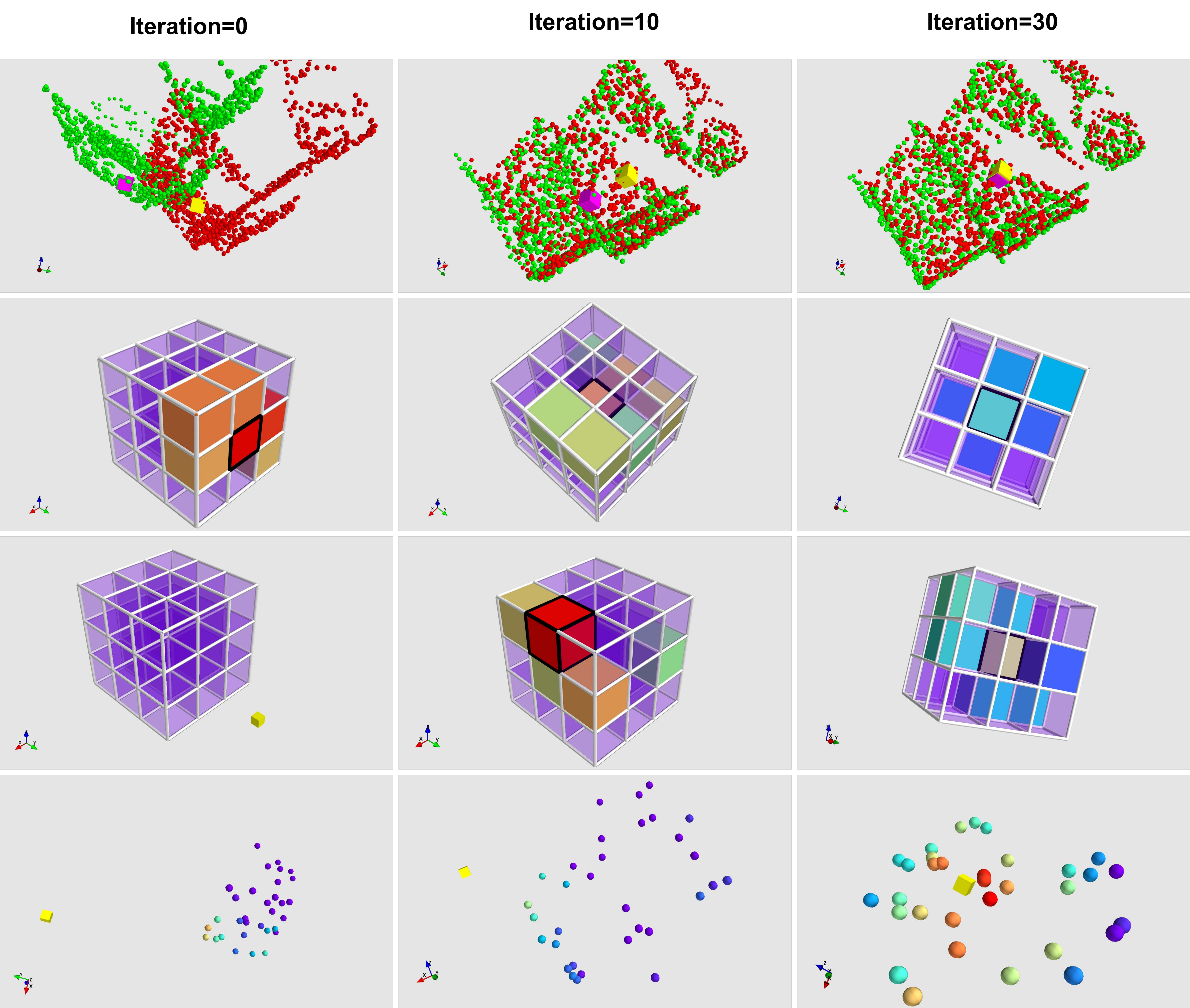}
	\caption{Visualization of point-voxel correlation fields. In the first row, green points represent translated point cloud $P_1$ + $\textbf{f}$ while red points stand for target point cloud $P_2$. The pink cube is a point in the translated point cloud, whose correspondence in $P_2$ is the yellow cube. The correlation fields of voxel branch are illustrated in the second ($r=1$) and third ($r=0.25$) rows. If the target point (yellow cube) lies in a lattice, the boundaries of this lattice will be colored in black. The last row exhibits the correlation field of the point branch. The colors of the last three rows indicate normalized correlation scores, where red is highest and purple is lowest (Figure \ref{fig:overview} shows colormap).  \textbf{At the beginning of the iterative update} (the first column), the predicted flow is not accurate so that the translated point is far from the target point. Since the voxel branch has large receptive fields, it can cover the target point while the point branch fails. From the first column and the second row, we see that the sub-cube which contains the target point has the highest correlation score. This indicates that the voxel branch provides effective guidance for flow prediction at early iterations. \textbf{As the iteration goes on}, the translated point gets near to the target point (the third column). The voxel branch only provides the coarse position of the target point (at the central sub-cube) while the point branch can accurately localize the target point by computing correlation scores of all neighbor points in the local region. The viewpoints are chosen to best visualize the sub-cube with the highest score. }
	\label{fig:corr}
\end{figure*}

\subsection{Ablation Studies} \label{sec:ablation}
We conducted experiments to confirm the effectiveness of each module in our method. Point-based correlation, voxel-based correlation and refinement module were applied to our framework incrementally. From Table \ref{tab:ablation}, we can conclude that each module plays an important part in the whole pipeline. As two baselines, the methods with only point-based correlation or voxel-based correlation fail to achieve high performance, since they cannot capture all-pairs relations. An intuitive solution is to employ more nearest neighbors in the point branch to increase the receptive field or decrease the side length $r$ in the voxel branch to take fine-grained correlations. However, we find that such straightforward methods lead to inferior results (See details in the supplemental material).

To better illustrate the effects of two types of correlations, we show visualizations in Figure \ref{fig:corr}. At the beginning of update steps, when predicted flows are initialized as zero, the estimated translated points are far from ground-truth correspondences in the target point cloud (first column). Under this circumstance, the similarity scores with near neighbors are small, where point-based correlation provides invalid information. In contrast, since voxel-based correlation has the large receptive field, it is able to find long-range correspondences and guide the prediction direction. As the update iteration increases, we will get more and more accurate scene flow. When translated points are near to the ground-truth correspondences, high-score correlations will concentrate on the centered lattice of the voxel (third column), which does not serve detailed correlations. However, we will get informative correlations from the point branch since KNN perfectly encodes local information.

\subsection{Further Analysis}
\begin{table}[tb]\footnotesize
	\centering
	\caption{Effects of truncation operation. $M$ denotes the truncation number.}
	\resizebox{0.45\textwidth}{!}{
		\begin{tabular}{cc|ccc}
			\hline
			$M$ & memory & EPE(m)$\downarrow$ & Acc Strict$\uparrow$     & Outliers$\downarrow$     \\
			\hline
			128 & 7.4G
			& 0.0585    & 0.7113     & 0.3810       \\
			512 & 10.7G
			& \bf{0.0461} & 0.8169 &0.2924      \\
			1024 & 14.1G
			&0.0475 &\bf{0.8173} &\bf{0.2910}       \\
			\hline
		\end{tabular}
		
	}
	\label{tab:truncate}
\end{table}

\begin{table}[tb]\footnotesize
	\centering
	\caption{Comparison with other correlation volume methods. "MLP+Maxpool" and "patch-to-patch" are correlation volumes used in FlowNet3D \cite{liu2019flownet3d} and PointPWC-Net \cite{wupointpwc} respectively. }
	\resizebox{0.48\textwidth}{!}{
		\begin{tabular}{l|ccc}
			\hline
			Method & EPE(m)$\downarrow$ & Acc Strict$\uparrow$    & Outliers$\downarrow$     \\
			\hline
			MLP+Maxpool \cite{liu2019flownet3d}
			& 0.0704    & 0.7137     & 0.3843    \\
			patch-to-patch \cite{wupointpwc}
			& 0.0614    & 0.7209     & 0.3628        \\  \hline
			point-voxel
			&\bf{0.0461}&\bf{0.8169}&\bf{0.2924}        \\
			\hline
		\end{tabular}
		
	}
	\label{tab:corr}
	\vspace{-3mm}
\end{table}

\noindent \textbf{Effects of Truncation Operation:}
We introduce the truncation operation to reduce running memory while maintain the performance.  To prove this viewpoint, we conducted experiments with different truncation numbers $M$, which are shown in Table \ref{tab:truncate}. On the one hand, when $M$ is too small, the accuracy will degrade due to the lack of correlation information. On the other hand, achieving the comparable performance with $M=512$, the model adopting $M=1024$ needs about 14G running memory, which is not available on many GPU services (\eg RTX 2080 Ti). This result indicates that top 512 correlations are enough to accurately estimate scene flow with high efficiency.

\noindent \textbf{Comparison with Other Correlation Volumes:}
To further demonstrate the superiority of the proposed point-voxel correlation fields, we did comparison with correlation volume methods introduced in FlowNet3D \cite{liu2019flownet3d} and PointPWC-Net \cite{wupointpwc}. To fairly compare, we applied their correlation volumes in our framework to substitute point-voxel correlation fields. Evaluation results are shown in Table \ref{tab:corr}. Leveraging all-pairs relations, our point-voxel correlation module outperforms other correlation volume methods.

\section{Conclusion}

In this paper, we have proposed a PV-RAFT method for scene flow estimation of point clouds. With the point-voxel correlation fields, our method integrates two types of correlations and captures all-pairs relations. Leveraging the truncation operation and the refinement module, our framework becomes more accurate. Experimental results on the FlyingThings3D and KITTI datasets verify the superiority and generalization ability of PV-RAFT.

\section*{Acknowledgement}
This work was supported in part by the National Natural Science Foundation of China under Grant U1713214, Grant U1813218, Grant 61822603, in part by Beijing Academy of Artificial Intelligence (BAAI), and in part by a grant from the Institute for Guo Qiang, Tsinghua University.

{\small
\bibliographystyle{ieee_fullname}
\bibliography{egbib}

\begin{thebibliography}{10}\itemsep=-1pt

\bibitem{basha2013multi}
Tali Basha, Yael Moses, and Nahum Kiryati.
\newblock Multi-view scene flow estimation: A view centered variational
  approach.
\newblock {\em IJCV}, 2013.

\bibitem{vcech2011scene}
Jan {\v{C}}ech, Jordi Sanchez-Riera, and Radu Horaud.
\newblock Scene flow estimation by growing correspondence seeds.
\newblock In {\em CVPR}, 2011.

\bibitem{chang2015shapenet}
Angel~X Chang, Thomas Funkhouser, Leonidas Guibas, Pat Hanrahan, Qixing Huang,
  Zimo Li, Silvio Savarese, Manolis Savva, Shuran Song, Hao Su, et~al.
\newblock Shapenet: An information-rich 3d model repository.
\newblock {\em arXiv preprint arXiv:1512.03012}, 2015.

\bibitem{dewan2016rigid}
Ayush Dewan, Tim Caselitz, Gian~Diego Tipaldi, and Wolfram Burgard.
\newblock Rigid scene flow for 3d lidar scans.
\newblock In {\em IROS}, 2016.

\bibitem{dosovitskiy2015flownet}
Alexey Dosovitskiy, Philipp Fischer, Eddy Ilg, Philip Hausser, Caner Hazirbas,
  Vladimir Golkov, Patrick Van Der~Smagt, Daniel Cremers, and Thomas Brox.
\newblock Flownet: Learning optical flow with convolutional networks.
\newblock In {\em CVPR}, 2015.

\bibitem{gu2019hplflownet}
Xiuye Gu, Yijie Wang, Chongruo Wu, Yong~Jae Lee, and Panqu Wang.
\newblock Hplflownet: Hierarchical permutohedral lattice flownet for scene flow
  estimation on large-scale point clouds.
\newblock In {\em CVPR}, 2019.

\bibitem{hou20193d}
Ji Hou, Angela Dai, and Matthias Nie{\ss}ner.
\newblock 3d-sis: 3d semantic instance segmentation of rgb-d scans.
\newblock In {\em CVPR}, 2019.

\bibitem{huguet2007variational}
Fr{\'e}d{\'e}ric Huguet and Fr{\'e}d{\'e}ric Devernay.
\newblock A variational method for scene flow estimation from stereo sequences.
\newblock In {\em ICCV}, 2007.

\bibitem{hui2018liteflownet}
Tak-Wai Hui, Xiaoou Tang, and Chen Change~Loy.
\newblock Liteflownet: A lightweight convolutional neural network for optical
  flow estimation.
\newblock In {\em CVPR}, 2018.

\bibitem{hur2020self}
Junhwa Hur and Stefan Roth.
\newblock {Self-Supervised Monocular Scene Flow Estimation}.
\newblock In {\em CVPR}, 2020.

\bibitem{ilg2017flownet}
Eddy Ilg, Nikolaus Mayer, Tonmoy Saikia, Margret Keuper, Alexey Dosovitskiy,
  and Thomas Brox.
\newblock Flownet 2.0: Evolution of optical flow estimation with deep networks.
\newblock In {\em CVPR}, 2017.

\bibitem{jiang2020pointgroup}
Li Jiang, Hengshuang Zhao, Shaoshuai Shi, Shu Liu, Chi-Wing Fu, and Jiaya Jia.
\newblock {PointGroup}: {Dual-Set Point Grouping for 3D Instance Segmentation}.
\newblock In {\em CVPR}, 2020.

\bibitem{kingma2014adam}
Diederik~P Kingma and Jimmy Ba.
\newblock Adam: A method for stochastic optimization.
\newblock {\em arXiv preprint arXiv:1412.6980}, 2014.

\bibitem{klokov2017escape}
Roman Klokov and Victor Lempitsky.
\newblock Escape from cells: Deep kd-networks for the recognition of 3d point
  cloud models.
\newblock In {\em ICCV}, 2017.

\bibitem{li2018so}
Jiaxin Li, Ben~M Chen, and Gim Hee~Lee.
\newblock So-net: Self-organizing network for point cloud analysis.
\newblock In {\em CVPR}, 2018.

\bibitem{li2019stereo}
Peiliang Li, Xiaozhi Chen, and Shaojie Shen.
\newblock Stereo r-cnn based 3d object detection for autonomous driving.
\newblock In {\em CVPR}, 2019.

\bibitem{li2018pointcnn}
Yangyan Li, Rui Bu, Mingchao Sun, Wei Wu, Xinhan Di, and Baoquan Chen.
\newblock Pointcnn: Convolution on x-transformed points.
\newblock In {\em NeurIPS}, 2018.

\bibitem{liu2019flownet3d}
Xingyu Liu, Charles~R Qi, and Leonidas~J Guibas.
\newblock Flownet3d: Learning scene flow in 3d point clouds.
\newblock In {\em CVPR}, 2019.

\bibitem{liu2019point}
Zhijian Liu, Haotian Tang, Yujun Lin, and Song Han.
\newblock Point-voxel cnn for efficient 3d deep learning.
\newblock In {\em NeurIPS}, 2019.

\bibitem{mayer2016large}
Nikolaus Mayer, Eddy Ilg, Philip Hausser, Philipp Fischer, Daniel Cremers,
  Alexey Dosovitskiy, and Thomas Brox.
\newblock A large dataset to train convolutional networks for disparity,
  optical flow, and scene flow estimation.
\newblock In {\em CVPR}, 2016.

\bibitem{menze2015object}
Moritz Menze and Andreas Geiger.
\newblock Object scene flow for autonomous vehicles.
\newblock In {\em CVPR}, 2015.

\bibitem{menze2015joint}
Moritz Menze, Christian Heipke, and Andreas Geiger.
\newblock Joint 3d estimation of vehicles and scene flow.
\newblock {\em ISPRS Annals of Photogrammetry, Remote Sensing \& Spatial
  Information Sciences}, 2015.

\bibitem{mittal2020just}
Himangi Mittal, Brian Okorn, and David Held.
\newblock Just go with the flow: Self-supervised scene flow estimation.
\newblock In {\em CVPR}, 2020.

\bibitem{paszke2017automatic}
Adam Paszke, Sam Gross, Soumith Chintala, Gregory Chanan, Edward Yang, Zachary
  DeVito, Zeming Lin, Alban Desmaison, Luca Antiga, and Adam Lerer.
\newblock Automatic differentiation in pytorch.
\newblock 2017.

\bibitem{pons2007multi}
Jean-Philippe Pons, Renaud Keriven, and Olivier Faugeras.
\newblock Multi-view stereo reconstruction and scene flow estimation with a
  global image-based matching score.
\newblock {\em IJCV}, 2007.

\bibitem{puy2020flot}
Gilles Puy, Alexandre Boulch, and Renaud Marlet.
\newblock {FLOT: Scene Flow on Point Clouds guided by Optimal Transport}.
\newblock In {\em ECCV}, 2020.

\bibitem{qi2019deep}
Charles~R Qi, Or Litany, Kaiming He, and Leonidas~J Guibas.
\newblock Deep hough voting for 3d object detection in point clouds.
\newblock In {\em ICCV}, 2019.

\bibitem{qi2017pointnet}
Charles~R Qi, Hao Su, Kaichun Mo, and Leonidas~J Guibas.
\newblock Pointnet: Deep learning on point sets for 3d classification and
  segmentation.
\newblock In {\em CVPR}, 2017.

\bibitem{qi2017pointnet++}
Charles~Ruizhongtai Qi, Li Yi, Hao Su, and Leonidas~J Guibas.
\newblock Pointnet++: Deep hierarchical feature learning on point sets in a
  metric space.
\newblock In {\em NeurIPS}, 2017.

\bibitem{ranjan2017optical}
Anurag Ranjan and Michael~J Black.
\newblock Optical flow estimation using a spatial pyramid network.
\newblock In {\em CVPR}, 2017.

\bibitem{rao2020global}
Yongming Rao, Jiwen Lu, and Jie Zhou.
\newblock Global-local bidirectional reasoning for unsupervised representation
  learning of 3d point clouds.
\newblock In {\em CVPR}, 2020.

\bibitem{shi2019pv}
Shaoshuai Shi, Chaoxu Guo, Li Jiang, Zhe Wang, Jianping Shi, Xiaogang Wang, and
  Hongsheng Li.
\newblock {PV-RCNN: Point-Voxel Feature Set Abstraction for 3D Object
  Detection}.
\newblock In {\em CVPR}, 2020.

\bibitem{shi2019pointrcnn}
Shaoshuai Shi, Xiaogang Wang, and Hongsheng Li.
\newblock Pointrcnn: 3d object proposal generation and detection from point
  cloud.
\newblock In {\em CVPR}, 2019.

\bibitem{su2018splatnet}
Hang Su, Varun Jampani, Deqing Sun, Subhransu Maji, Evangelos Kalogerakis,
  Ming-Hsuan Yang, and Jan Kautz.
\newblock Splatnet: Sparse lattice networks for point cloud processing.
\newblock In {\em CVPR}, 2018.

\bibitem{sun2018pwc}
Deqing Sun, Xiaodong Yang, Ming-Yu Liu, and Jan Kautz.
\newblock Pwc-net: Cnns for optical flow using pyramid, warping, and cost
  volume.
\newblock In {\em CVPR}, 2018.

\bibitem{tang2020searching}
Haotian Tang, Zhijian Liu, Shengyu Zhao, Yujun Lin, Ji Lin, Hanrui Wang, and
  Song Han.
\newblock Searching efficient 3d architectures with sparse point-voxel
  convolution.
\newblock In {\em ECCV}, 2020.

\bibitem{teed2020raft}
Zachary Teed and Jia Deng.
\newblock {RAFT: Recurrent All-Pairs Field Transforms for Optical Flow}.
\newblock In {\em ECCV}, 2020.

\bibitem{truong2020glu}
Prune Truong, Martin Danelljan, and Radu Timofte.
\newblock {GLU-Net: Global-Local Universal Network for Dense Flow and
  Correspondences}.
\newblock In {\em CVPR}, 2020.

\bibitem{ushani2018feature}
Arash~K Ushani and Ryan~M Eustice.
\newblock Feature learning for scene flow estimation from lidar.
\newblock In {\em Conference on Robot Learning}, 2018.

\bibitem{ushani2017learning}
Arash~K Ushani, Ryan~W Wolcott, Jeffrey~M Walls, and Ryan~M Eustice.
\newblock A learning approach for real-time temporal scene flow estimation from
  lidar data.
\newblock In {\em ICRA}, 2017.

\bibitem{vedula2005three}
Sundar Vedula, Peter Rander, Robert Collins, and Takeo Kanade.
\newblock Three-dimensional scene flow.
\newblock {\em IEEE TPAMI}, 2005.

\bibitem{vogel20113d}
Christoph Vogel, Konrad Schindler, and Stefan Roth.
\newblock 3d scene flow estimation with a rigid motion prior.
\newblock In {\em ICCV}, 2011.

\bibitem{vogel2013piecewise}
Christoph Vogel, Konrad Schindler, and Stefan Roth.
\newblock Piecewise rigid scene flow.
\newblock In {\em CVPR}, 2013.

\bibitem{vogel20153d}
Christoph Vogel, Konrad Schindler, and Stefan Roth.
\newblock 3d scene flow estimation with a piecewise rigid scene model.
\newblock {\em IJCV}, 2015.

\bibitem{wang2018sgpn}
Weiyue Wang, Ronald Yu, Qiangui Huang, and Ulrich Neumann.
\newblock Sgpn: Similarity group proposal network for 3d point cloud instance
  segmentation.
\newblock In {\em CVPR}, 2018.

\bibitem{wang2019dynamic}
Yue Wang, Yongbin Sun, Ziwei Liu, Sanjay~E Sarma, Michael~M Bronstein, and
  Justin~M Solomon.
\newblock Dynamic graph cnn for learning on point clouds.
\newblock {\em TOG}, 2019.

\bibitem{wedel2011stereoscopic}
Andreas Wedel, Thomas Brox, Tobi Vaudrey, Clemens Rabe, Uwe Franke, and Daniel
  Cremers.
\newblock Stereoscopic scene flow computation for 3d motion understanding.
\newblock {\em IJCV}, 2011.

\bibitem{wedel2008efficient}
Andreas Wedel, Clemens Rabe, Tobi Vaudrey, Thomas Brox, Uwe Franke, and Daniel
  Cremers.
\newblock Efficient dense scene flow from sparse or dense stereo data.
\newblock In {\em ECCV}, 2008.

\bibitem{wei2019conditional}
Yi Wei, Shaohui Liu, Wang Zhao, and Jiwen Lu.
\newblock Conditional single-view shape generation for multi-view stereo
  reconstruction.
\newblock In {\em CVPR}, 2019.

\bibitem{wupointpwc}
Wenxuan Wu, Zhi~Yuan Wang, Zhuwen Li, Wei Liu, and Li Fuxin.
\newblock {PointPWC-Net: Cost Volume on Point Clouds for (Self-) Supervised
  Scene Flow Estimation}.
\newblock In {\em ECCV}, 2020.

\bibitem{yang20203dssd}
Zetong Yang, Yanan Sun, Shu Liu, and Jiaya Jia.
\newblock {3dssd: Point-based 3d single stage object detector}.
\newblock In {\em CVPR}, 2020.

\end{thebibliography}
}
\clearpage

\twocolumn[{
	\vspace{-3em}
\begin{center}
	\centering
	\captionof{table}{The necessity of point-voxel correlation fields. We conducted experiments on FlyingThings3D dataset without refinement. KNN pyramid means we concatenated correlation features with different $K$. }
	\resizebox{0.9\textwidth}{!}{
		\begin{tabular}{l|l|cccc}
			\hline
			 Modality &Hyperparameters &EPE(m)$\downarrow$   & Acc Strict$\uparrow$ & Acc Relax$\uparrow$& Outliers$\downarrow$     \\ 
			\hline
		     \multirow{3}{*}{KNN} &  $K=32$    
			& 0.0741    & 0.6111    & 0.8868    & 0.4549       \\
			 &  $K=64$  
			& 0.2307    & 0.1172    & 0.3882    & 0.8547       \\
		    &  $K=128$    
			& 0.6076    & 0.0046    & 0.0333   & 0.9979       \\ \hline
			\multirow{2}{*}{KNN pyramid} &  $K=16,32,64$     
			& 0.1616    & 0.2357   & 0.6062    & 0.7318       \\
			 &  $K=32,64,128$    
			& 0.4841    & 0.0158    & 0.0885    & 0.9882      \\ \hline
			\multirow{4}{*}{voxel pyramid}& $r=0.0625, l=3$   
			&  0.1408	&0.5126	    &0.8057	    &0.5340      \\
			& $r=0.125, l=3$  
			&  0.0902	&0.5345	    &0.8533	    &0.5085      \\
			& $r=0.25, l=3$   
			&  0.0712	&0.6146	    &0.8983	    &0.4492      \\
			& $r=0.0625, l=5$   
			&  0.0672	&0.6325	    &0.9131    &0.4023      \\ \hline
			
			point-voxel &$K=32,r=0.25,l=3$&\bf{0.0534}    & \bf{0.7348}    & \bf{0.9418}    & \bf{0.3645}        \\
			 \hline
		\end{tabular}
		
	}
	\label{tab:exp}
\end{center}
}]
\section*{Appendix}
\appendix
\section{Network Architecture}
The architecture of our network can be divided into four parts: (1) Feature Extractor, (2) Correlation Module (3) Iterative Update Module (4) Refinement Module. In this section, we will introduce the implementation details of each structure. 

\subsection{Feature Extractor}
\noindent \textbf{Backbone Feature Extractor}
We first construct a graph $\mathcal{G}$ of input point cloud $P$, that contains neighborhood information of each point. Then we follow FLOT which is based on PointNet++ to design the feature extractor.

The feature extractor consists of three SetConvs to lift feature dimension: $3 \to 32 \to 64 \to 128$. In each SetConv, we first locate neighbor region $\mathcal{N}$ of $P$ and use $F = concat(F_{\mathcal{N}} - F_P, F_{\mathcal{N}})$ as input features, where $concat$ stands for concatenation operation. Then features $F$ are fed into the pipeline: $FC \to pool \to FC \to FC$. Each $FC$ block consists of a 2D convolutional layer, a group normalization layer and a leaky ReLU layer with the negative slope as $0.1$. If we denote the input and output dimension of the SetConv as $d_i, d_o$, then the dimension change for $FC$ blocks is: $d_i \to d_{mid}=(d_i + d_o)/2 \to d_o \to d_o$. However, if $d_i=3$, then $d_{mid}$ is set to $d_o/2$. The $pool$ block performs the max-pooling operation. 

\noindent \textbf{Context Feature Extractor}
The context feature extractor aims to encode context features of $P_1$. It has exactly the same structure as the backbone feature extractor, but without weight sharing.

\subsection{Correlation Module}
\noindent \textbf{Point Branch}
The extracted KNN features $F_{p}(P)$ are first concatenated with position features $C({\mathcal{N}_P}) - C(P)$, then it is fed into a block that consists of one point-wise convolutional layer, one group normalization layer, one p-ReLU layer, one max-pooling layer and one point-wise convolutional layer. The feature dimension is updated from $4$ to $64$.

\noindent \textbf{Voxel Branch}
The extracted voxel features $F_{v}(P)$ are fed into a block that consists of one point-wise convolutional layer, one group-norm layer, one p-ReLU layer and one point-wise convolutional layer. The feature dimension is updated as: $a^3 * l \to 128 \to 64$, where $a=3$ is the resolution hyper-parameter and $l=3$ is the pyramid level.

\subsection{Iterative Update Module}
The update block consists of three parts: Motion Encoder, GRU Module and Flow Head.

\noindent \textbf{Motion Encoder}
The inputs of motion encoder are flow $f$ and correlation features $\textbf{C}$. These two inputs are first fed into a non-share convolutional layer and a ReLU layer separately to get $f'$ and $\textbf{C}'$. Then they are concatenated and fed into another convolutional layer and a ReLU layer to get $f''$. Finally we concat $f$ and $f''$ to get motion features $f_m$.

\noindent \textbf{GRU Module}
The inputs of GRU module are context features and motion features. The update process has already been introduced in our main paper.

\noindent \textbf{Flow Head}
The input of the flow head is the final hidden state $h_t$ of GRU module. $h_t$ is first fed into a 2D convolutional layer to get $h'_t$. On the other hand, $h_t$ is fed into a SetConv layer, introduced in backbone feature extractor, to get $h''_t$. Then we concatenate $h'_t$ and $h''_t$ and pass through a 2D convolutional layer to adjust the feature dimension to $3$. The output is used to update flow prediction. 

\subsection{Refinement Module}
The input of the refinement module is the predicted flow $f^*$. The refinement module consists of three SetConv modules and one Fully Connected Layer. The SetConv module has been introduced in feature extractor part and the dimension is changed as: $3 \to 32 \to 64 \to 128$. The output feature $f^*_r$ of fully connected layer is of dimension $3$. We implement a residual mechanism to get the final prediction that combines $f^*$ and $f^*_r$. 

\section{Additional Experiments} \label{sec:experiment}
As mentioned in Section 4.3, we tried intuitive solutions to model all-pairs correlations. We conducted experiments on FlyingThings3D dataset without refinement. Specifically, for the point branch, we leveraged more nearest neighbors to encode large receptive fields. When only using the voxel branch, we reduce the side length $r$ of lattices to capture fine-grained relations. Moreover, we adopted the KNN search with different $K$ simultaneously to construct a KNN pyramid , which aims to aggregate the feature with different receptive fields. However, as shown in Table \ref{tab:exp}, all these tries failed to achieve promising results. We argue that this may because of the irregularity of point clouds. On the one hand, for the region with high point density, a large number of neighbors still lead to a small receptive field. On the other hand, although we reduce side length, the voxel branch cannot extract point-wise correlation features. Integrating these two types of correlations, the proposed point-voxel correlation fields help PV-RAFT to capture both local and long-range dependencies. 
\end{document}